\documentclass[11pt,a4paper]{article}
\usepackage[hyperref]{acl2020}
\usepackage{times}
\usepackage{latexsym}

\usepackage{microtype}

\usepackage[colorinlistoftodos, textsize=tiny]{todonotes}
\usepackage{amsfonts}
\usepackage{mathtools}
\usepackage{qtree}
\usepackage{hhline}
\usepackage{multirow}
\DeclareMathOperator*{\argmax}{arg\,max}

\usepackage{url}

\aclfinalcopy 

\mathtoolsset{showonlyrefs,showmanualtags}

\title{Obfuscation for Privacy-preserving Syntactic Parsing}

\author{Zhifeng Hu$^{\spadesuit\ast}$ \quad
  Serhii Havrylov$^{\diamondsuit}$ \quad Ivan Titov$^{\diamondsuit}{^\heartsuit}$ \quad Shay B. Cohen$^{\diamondsuit}$
 \smallskip \\
$^\spadesuit$School of Computer Science, Fudan University, Shanghai 201203, China \\
$^\diamondsuit$School of Informatics, University of Edinburgh, Edinburgh EH8 9AB, UK \\
$^\heartsuit$ILLC / FNWI, University of Amsterdam, Amsterdam 1098XG, Netherlands \\
{\tt  zfhu16@gmail.com}, {\tt s.havrylov@ed.ac.uk } \\ {\tt  ititov@inf.ed.ac.uk},
{\tt scohen@inf.ed.ac.uk}
}

\newcommand\blfootnote[1]{%
  \begingroup
  \renewcommand\thefootnote{}\footnote{#1}%
  \addtocounter{footnote}{-1}%
  \endgroup
}

\date{}

\newcommand{\ignore}[1]{}
\newcommand{\shaycomment}[1]{\textcolor{blue}{#1}}
\newcommand{\zhifengcomment}[1]{\textcolor{purple}{#1}}

\begin{document}
\maketitle
\begin{abstract}
The goal of homomorphic encryption is to encrypt data such that another party can operate on it without being explicitly exposed to the content of the original data.
We introduce an idea for a privacy-preserving transformation on natural language data, inspired by homomorphic encryption.
Our primary tool is {\em obfuscation}, relying on the properties of natural language.
Specifically, a given English text is obfuscated using a neural model that aims to preserve the syntactic relationships of the original sentence so that the obfuscated sentence
can be parsed instead of the original one.
The model works at the word level, and learns to obfuscate each word separately by changing it into a new word that has a similar syntactic role. 
The text obfuscated by our model leads to better performance on three syntactic parsers (two dependency and one constituency parsers) in comparison to an upper-bound random substitution baseline.
More specifically, the results demonstrate that as more terms are obfuscated (by their part of speech), the substitution upper bound significantly degrades, while the neural model maintains
a relatively high performing parser. All of this is done without much sacrifice of privacy compared to the random substitution upper bound.
We also further analyze the results, and discover that the substituted words have similar syntactic properties, but different semantic content, compared to the original words.
\end{abstract}

\section{Introduction}

We consider the case in which there is a powerful server with NLP technology deployed on it, and a set of clients who would like to access it to get output resulting from input text taken from problems such as syntactic parsing, semantic parsing and machine translation.\blfootnote{$^\ast$ Work done at the University of Edinburgh.}
In such a case, the server models may have been trained on large amounts of data, yielding models that cannot be deployed on the client machines either for efficiency or licensing reasons.
We ask the following question: how can we use the NLP server models while minimizing the exposure of the server to the original text? Can we exploit the fact we work with natural language data to reduce such exposure?


\begin{figure}[t]
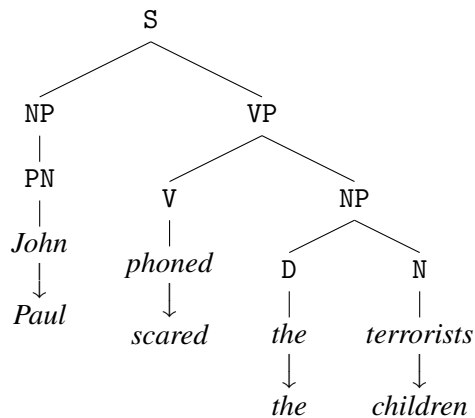


\begin{center}
\Tree [.${\tt S}$ [.${\tt NP}$ [.${\tt PN}$ $\begin{array}{c}\textit{John}\\\big\downarrow\\\textit{Paul}\end{array}$ ] ]
[.${\tt VP}$ [.${\tt V}$ $\begin{array}{c}\textit{phoned}\\\big\downarrow\\\textit{scared}\end{array}$ ] [.${\tt NP}$ [.${\tt D}$ $\begin{array}{c}\textit{the}\\\big\downarrow\\\textit{the}\end{array}$ ] [.${\tt N}$ $\begin{array}{c}\textit{terrorists}\\\big\downarrow\\\textit{children}\end{array}$ ] ] ] ]
\end{center}
\caption{\label{fig:obfuscate} An example of a sentence (words on top) and an obfuscated version of the sentence (words at bottom), both having identical syntactic structure.
The obfuscated sentence hides the identity of the person who performs the action and the action itself.
}
\end{figure}

Conventional encryption schemes, including public-key cryptography which is the one widely used across the Internet, are not sufficient to answer this question.
They encrypt the input text before it is transferred to the server side.
However, once the server decrypts the text, it has full access to it.
This might be unacceptable if the server itself is not necessarily trustworthy.

The cryptography community posed a similar question much earlier, in the 1970s \cite{rivest1978data} with partial resolutions proposed to solve it in later research \cite{sander1999non,boneh2005evaluating,ishai2007evaluating}.
These solutions allow the server to perform computations directly on encrypted data to get the desired output without ever decrypting the data.
This cryptographic protocol is known as {\em homomorphic encryption}, where a client encrypts a message, then sends it to a server which performs potentially computationally intensive operations and returns a new data, still encrypted, which only the client can decipher.
All of this is done without the server itself ever being exposed to the actual content of the encrypted input data.
While solutions for generic homomorphic encryption have been discovered, they are either computationally inefficient \cite{DBLP:journals/cacm/Gentry10} or have strong limitations in regards to the depth and complexity of computation they permit \cite{DBLP:conf/ima/BosLLN13}.

In this paper, we consider a softer version of homomorphic encryption in the form of {\em obfuscation} for natural language.
Our goal is to identify an efficient function that stochastically transforms a given natural language input (such as a sentence) into another input which can be further fed into an NLP server.
The altered input has to preserve intra-text relationships that exist in the original sentence such that the NLP server, depending on the task at hand, can be successfully applied on the transformed data.
There should be then a simple transformation that maps the output on the obfuscated data into a valid, accurate output for the original input.
In addition, the altered input should hide the private semantic content of the original data.

This idea is demonstrated in Figure~\ref{fig:obfuscate}. The task at hand is syntactic parsing.
We transform the input sentence \emph{John phoned the terrorists} to the sentence \emph{Paul scared the children} -- both of which yield identical phrase-structure trees.
In this case, the named entity \emph{John} is hidden, and so are his actions. 
In the rest of the paper, we focus on this problem for dependency and constituency parsing.

We consider a neural model of obfuscation that operates at the word level. 
We assume access to the parser at training time: the model learns how to substitute words in the sentence with other words (in a stochastic manner) while maintaining the highest possible parsing accuracy.
This learning task is framed as a latent-variable modeling problem where the obfuscated words are treated as latent.
Direct optimization of this model turns out to be intractable, so we use continuous relaxations \cite{DBLP:journals/corr/JangGP16,maddison2016concrete}
to avoid explicit marginalization.

Our experimental results on English demonstrate that the neural model performs better than a strong random-based baseline (an upper bound; in which a word is substituted randomly with another word with the same part-of-speech tag).
We vary the subset of words that are hidden and observe that the higher the obfuscation rate of the words, the harder it becomes for the parser to retain its accuracy. Degradation is especially pronounced with the random baseline and is less severe with our neural
model. The improved results for the neural obfuscator come at a small cost to the accuracy of the attacker aimed at recovering the original obfuscated words. We also observe that the neural obfuscator is effective when different parsers or even different syntactic formalisms are used in training and test time. This relaxes the assumption that the obfuscator needs to have access to the NLP server at training time.
Our results also suggest that the neural model tends to replace words
with ones that have similar syntactic properties.

\begin{figure}
\begin{center}
    \includegraphics[width=2in]{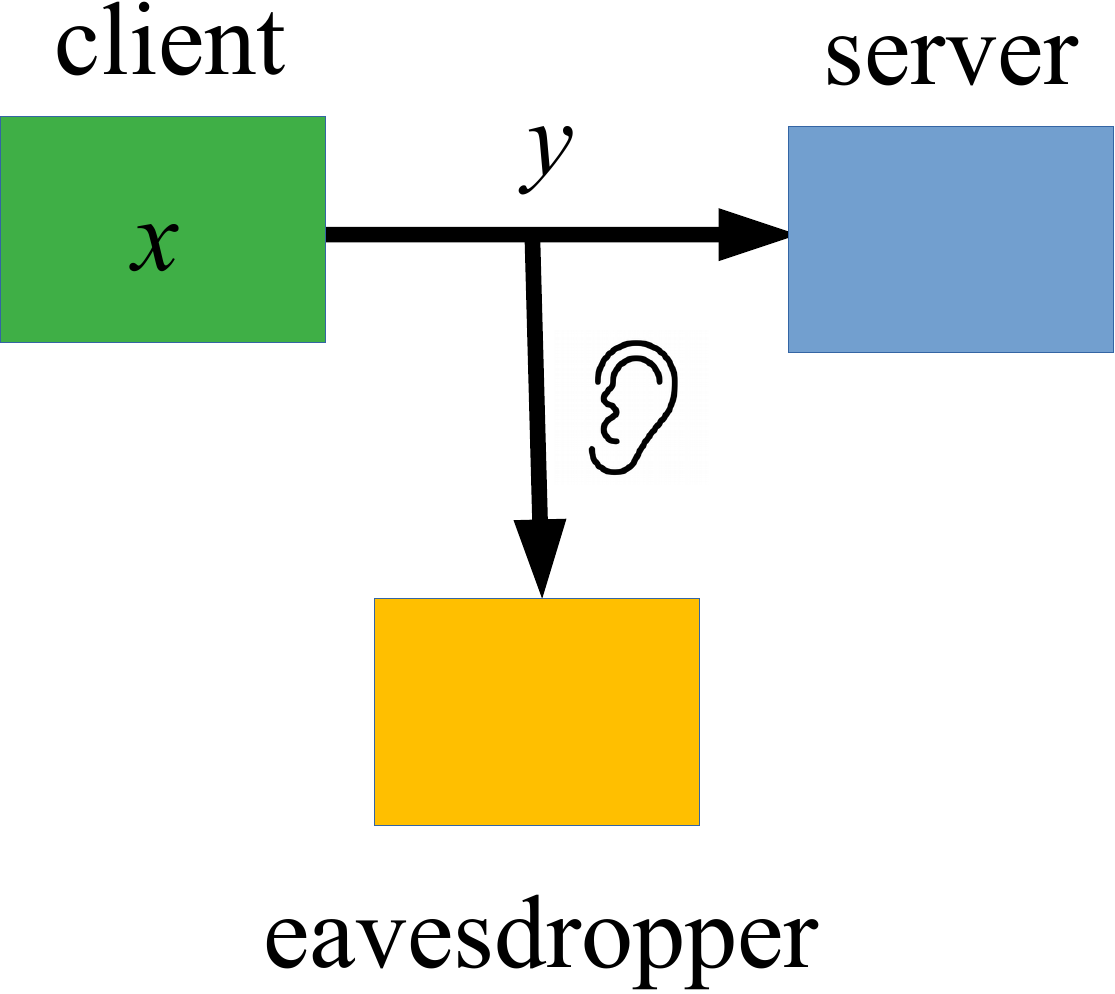}
\end{center}
    \caption{General setting illustration (figure adapted from \citealt{DBLP:conf/emnlp/CoavouxNC18}). An NLP client encrypts an $x$ into $y$ through obfuscation and $y$ is sent to an NLP server. 
    The NLP server (potentially even a legacy one) does not need to be modified to de-obfuscate $y$. An eavesdropper (a possibly malicious channel listener) only has access to $y$ which is needed to be de-obfuscated to gain any information about $x$.}
    \label{fig:setting}
\end{figure}

\section{Homomorphic Obfuscation of Text} 

Our problem formulation is rather simple, demonstrated in generality in Figure~\ref{fig:setting}. Let $\mathcal{T}$ be some natural language task, such as syntactic parsing, where $\mathcal{X}$ is the input space and
$\mathcal{Z}$ is the output space.
Let $f_{\mathcal{T}} \colon \mathcal{X} \rightarrow \mathcal{Z}$ be a trained decoder that maps $x$ to its corresponding structure according to $\mathcal{T}$. Note that $f$ is trained as usual on labeled data.
Given a sentence $x = x_1 \cdots x_n$, we aim to learn a function that stochastically transforms
$x$ into $y = y_1 \cdots y_n$ such that $f_{\mathcal{T}}(x)$ is close, if not identical, to $f_{\mathcal{T}}(y)$, or at the very least, we would like to be able
to recover $f_{\mathcal{T}}(x)$ from $f_{\mathcal{T}}(y)$ using a simple transformation.

To ground this in an example, consider the case in which $\mathcal{T}$ is the problem of dependency parsing and $\mathcal{Z}$ is the set of dependency trees.
If we transform a sentence $x$ to $y$ in such a way that it preserves the syntactic relationship between the indexed words in the sentences, then we can expect to easily recover the dependency tree for $x$ from a dependency tree for $y$.

Note that we would also want to stochastically transform $x$ into a $y$ in such a way that it is {\em hard} to recover a certain type of information in $x$ from $y$ (otherwise, we could just set $y \leftarrow x$). 
Furthermore, we are interested in hiding information such as named entities or even nouns and verbs.
In our formulation, we also assume that the sentence $x$ comes with a function $t(x)$ that maps each token in the sentence with its corresponding
part-of-speech tag (predicted using a POS tagger). 

\section{Neural Obfuscation Model}

In this section we describe the neural model used to obfuscate the sentence. 
We note that the model has to be simple and efficient, as it is being run
by the obfuscating party. If it is more complicated than parsing the text, for example, then the obfuscating party might as well directly parse the
text.\footnote{In the general case, there is a caveat to this statement. It might be the case that the training cost for the server's model is high, and
that the model is proprietary. In that case, even if the model can be run on the client side, it might not be possible to do so.}

\subsection{The Main Model}
\label{section:main-model}

Our model operates by transforming a subset of the words in the sentence into new words. Each of these words is separately transformed in a way that maintains the sentence length after the transformation. Let $x$ be the original sentence $x = x_1 \cdots x_n$ and let $y$ be the output, $y = y_1 \cdots y_n$.
From a high-level point of view, we have a conditional model:
\begin{equation}
p(y \mid x, \theta) = \prod_{i=1}^n p(y_i \mid x, \theta). \label{eq:a}
\end{equation}

The selection of words to obfuscate depends on their part of speech (POS) tags -- only words that are associated with specific POS tags from the set $\mathcal{P}$ are obfuscated under our model.
Let $t_i$ be the POS tag of the $i$th word in the sentence.
In our basic model, we apply a bidirectional Long Short-Term Memory network (BiLSTM) to the sentence to get a latent representation $h_i$ for each word $x_i$ (see
Section~\ref{section:embedding}).

We assume conditional independence between the  sequence $x_1 \cdots x_{i-1} x_{i+1} \cdots x_n$ and $y_i$ given $h_i$ (which is a function of $x$), and as such,
our probability distribution $p(y_i \mid x, \theta)$ is given by:

\begin{align}
p(y_i = y  \mid &  x_i,  h_i, \theta) = \\
&
\begin{cases}
1 &\! t_i \notin \mathcal{P}, y = x_i \\
p_y & t_i \in \mathcal{P}\!,  y \in  \mathcal{V}_{t_i} \setminus \{x_i\} \\
0 &\! \textit{otherwise}.
\end{cases}
\label{eq:b}
\end{align}
Here, $\mathcal{V}_{t_i}$ is the set of word types appearing at least once with tag $t_i$ in the training set, and $p_y$ is predicted with a softmax function, relying on the BiLSTM state $h_i$.
More specifically, we define $p_y$ as follows:
\begin{equation}
p_y = 
\frac 
{\exp (w_{t_i,y}^{\top} h_i)} 
{\sum_{y' \in \mathcal{V}_{t_i}, y' \neq x_i}{\exp(w_{t_i,y'}^{\top} h_i)}},
\end{equation}
where  $w_{t,y} \in \mathbb{R}^{1024}$ are vectors of parameters associated with every tag-word pair $(t,y)$, $y \in \mathcal{V}_t$.  Note that the above probability distribution never transforms a word $x_i$ to an identical word if $t_i \in \mathcal{P}$.
This is a hard constraint in our model.



\subsection{Embedding the Sentence}
\label{section:embedding}

The BiLSTM that encodes the sentence requires an embedding per word, which we create as follows. We first map each token $x_i$
to three embedding channels \(e^k_i\), \(k \in \{1,2,3\}\).
The first channel is a randomly initialized embedding for each part-of-speech tag.
Its dimension is $100$. The second channel is a pre-trained GloVe embedding for the corresponding token.
The vector $e^3_i$ is a character-level word embedding \cite{DBLP:conf/aaai/KimJSR16} which first maps each character of the word into an embedding vector of dimension $100$ and then uses unidimensional convolution over the concatenation of the embedding vectors of each character.
Finally, max-pooling is applied to obtain a single feature. This process is repeated with $100$ convolutional kernels so that $e^3_i \in \mathbb{R}^{100}$.

The three embedding channels \(\{e^1_i,e^2_i,e^3_i\}\) are then concatenated and used in the BiLSTM encoder. We use a three-layer BiLSTM
with Bayesian dropout~\cite{DBLP:conf/nips/GalG16}. The hidden state dimensionality is 512 for each direction.

\section{Training}
\label{section:training}

In our experiments, we focus on obfuscation for the goal of syntactic parsing.
We assume the existence of a conditional parsing model $p_0(z \mid x)$ where $z$ is a parse tree and $x$ is a sentence.
This is the base model which is trained offline, and to which we have read-only access and cannot change its parameters.
As we will see in experiments, the obfuscator can be trained using a different parser from the one used at test time (i.e. from the one hosted at the NLP server).


Let $(x^{(1)}, z^{(1)}),\ldots,(x^{(n)}, z^{(n)})$ be a set of training examples which consists of sentences and their corresponding
parse trees. 
Considering Eq.~\refeq{eq:a}, we would be interested in maximizing the following log-likelihood objective with
respect to $\theta$:

\begin{equation}
\nonumber
\mathcal{L}_0 = 
\sum_{i=1}^n \log \left( \sum_y p(y \mid x^{(i)}, \theta) p_0(z^{(i)} \mid y) \right). 
\end{equation}

This objective maximizes the log-likelihood of the parsing model with respect to the obfuscation
model.
Maximizing the objective $\mathcal{L}_0$ is intractable due to summation over all possible obfuscations. We use Jensen's inequality\footnote{Jensen's inequality states that for
a non-negative random variable $Z$ and its probability distribution $q$ it holds that $\log (\mathbb{E}_q [ Z ]) \ge \mathbb{E}_q [\log Z]$.} 
to lower-bound the cost function $\mathcal{L}_0$ by the following objective:

\begin{align}
\mathcal{L} & = 
\sum_{i=1}^n \sum_y p(y \mid x^{(i)}, \theta) \log p_0(z^{(i)} \mid y) \\
	    & = 
	    \sum_{i=1}^n \mathbb{E}_{p(\cdot \mid x^{(i)}, \theta)}\left[ \log p_0(z^{(i)} \mid y)\right]. \label{eq:dd}
\end{align}

Intuitively, the objective function  maximizes the accuracy of an existing parser
while using as an input the sentences after their transformation. Note that the accuracy is measured with respect
to the gold-standard dependency parse tree.\footnote{In principle, we may not need access to gold-standard annotation when training the obfuscator. Instead, we could 
train the model to agree with the parser predictions for the original sentence, i.e. $z^{(i)} = \argmax_z p_0(z | x^{(i)})$.}
This is possible because the sentence length of the original sentence and
the obfuscated sentence are identical, and the mapping between the words in each version of the sentence is bijective.

To encourage stochasticity, we also tried including an entropy term that is maximized with respect to $\theta$ in the following form:
\begin{equation}
H_i(\theta, \lambda) =  - \lambda \sum_y p(y \mid x^{(i)}, \theta) \log p(y \mid x^{(i)}, \theta) .
\end{equation}
However, in our final experiments we omitted that term because (a) it did not seem to affect the model stochasticity to a significant degree; (b) the performance has become very sensitive to the entropy weight $\lambda$.

While we can estimate the objective $\mathcal{L}$ using sampling, we cannot differentiate through samples to estimate the gradients with respect to the obfuscator parameters $\theta$.
In order to ensure end-to-end differentiabilty, we use a continuous relaxation, the Gumbel-Softmax estimator~\cite{DBLP:journals/corr/JangGP16,maddison2016concrete}, and the reparamterization trick~\cite{kingma2013auto,rezende2014stochastic}.


More formally, the $i$-th token is represented by the random variable with categorical probability distribution \(\text{Cat}(p_i)\) that has support $\mathcal{V}_{t_i}$.
To sample the word we first draw $u_k\sim \mathrm{Uniform}(0,1)$ and transform it to the Gumbel noise $g_k=-\log(-\log(u_k))$, then we calculate
$$
y'=\mathrm{onehot}\left\{\argmax_{k \in \mathcal{V}_{t_i}} [g_k+\log(p_{i,k})]\right\}
$$
as the sampled discrete choice of substitution from $\mathcal{V}_{t_i}$ and 
$$
y_k=\frac{\exp\left((g_k+\log(p_{i,k})/\tau)\right)}{\sum_{k'} \exp\left((g_{k'}+\log(p_{i,k'})/\tau)\right)}
$$ 
as the ``relaxed'' differentiable proxy for this choice, where $\tau$ denotes the temperature. When it approaches $0$, the vector $(y_1, \ldots, y_{|\mathcal{V}_{t_i}|})$ 
is close to a one-hot vector sampled from the given categorical distribution.\footnote{In practice, we anneal the temperature from 1.0 to 0.5 over the course of training.}

We use the Straight-Through version of the estimator~\cite{bengio2013estimating}: the discrete sampled choice is fed into the parser in the forward computation but the relaxed differentiable surrogate is used when computing partial derivatives on the backward pass.

During the training of our neural model, the parser only backpropagates the gradient from the objective of maximizing the parsing accuracy (i.e. minimum cross-entropy loss of the correct head and label for each word), and hence its parameters are always fixed and are not updated during the optimization.

\section{Attacker Approaches}
\label{section:attack}

We test the efficiency of our obfuscation model by developing two independent attacker models. Their goal is to recover the original words by inspecting
only the obfuscated sentence. The attacker models may have access to all data that the parser and the obfuscator models were trained and developed on. This is perhaps unlike other
scenarios in which the training set is assumed to be inaccessible to any attacker.

We note that ideally, we would want to show that our obfuscation model retains privacy universally for {\em any} attacker. However, this is quite a difficult
task, and we follow \newcite{DBLP:conf/emnlp/CoavouxNC18} in presenting two strong attackers which may represent possible universal attackers.

In our attacker experiments, we assume that it is known which words in the sentence are obfuscated. As such, the results we provide for attacking our obfuscation
are an upper bound. In practice, an attacker would also have to identify which words were substituted for new words, which may lead to a small decrease in its accuracy.

\subsection{Trained Attacker}
\label{section:ff-attack}


Our first attacker works by first encoding the obfuscated sentence with a BiLSTM network. We then try to predict original words by using a feed-forward neural network on each of the hidden representations obtained from the encoder model. 
The architecture is identical to that of the obfuscation model (see Section~\ref{section:main-model}), with the only difference that there is a softmax over the entire vocabulary $\mathcal{V}$  instead of restricting it to $V_{t_i} \setminus \{x_i\}$, as in Eq.~\refeq{eq:b}.

\ignore{
More specifically, let $u_i$ be the predicted word for the $i$th word in the obfuscated sentence $y = y_1 \cdots y_n$. In addition, let $g_i$ be the latent representation
of the $i$th word in the obfuscated sentence with the LSTM encoder.
Then, we set $u_i$ to be the result of a softmax function as follows:

\begin{equation}
u_i = \mathrm{SoftMax}(W_2\mathrm{ReLU}(W_1g_i+b_1)+b_2).
\end{equation}

To identify the parameters $W_1, W_2, b_1, b_2$ we train the feed-forward neural network using cross-entropy loss.
}


\subsection{Pretrained Attacker}
\label{section:bert-attack}

In addition to a trained attacker, we also use a conditional language model, BERT \cite{devlin2018bert}.\footnote{We use the implementation available at
\url{https://github.com/huggingface/pytorch-pretrained-BERT}.} BERT is based on the Transformer model of \newcite{vaswani2017attention}, and uses a bidirectional encoder to
obtain ``contextual'' embeddings for each word in a given sentence.
We use the BERT model by masking out each obfuscated word, and then predicting the masked word similar to the ``masked language task'' that is mentioned
by \newcite{devlin2018bert}. This means that the embeddings in each position are fed into a softmax function to predict
the missing word. We use the {\tt bert-base-uncased} model among the available BERT models.

We note that this attacker is not trained by us. Its main weakness is that it is trained on the non-obfuscated text. However, its strength is that it is trained on large amounts of data (we use the model that is trained on 3.3 billion tokens).
In addition, in some settings that we consider the obfuscation of the sentence is done in such a way that much of the context by which we predict the obfuscated word remains intact.

\section{Experiments}
\label{section:experiments}
In this section, we describe our experiments with our obfuscation model. We first describe the experimental setting and then
turn to the results.\footnote{Our code is available at \url{https://github.com/ichn-hu/Parsing-Obfuscation}.}

\subsection{Experimental Setting}
\label{section:experimental-setup}
\label{section:setup}

In our experiments, we test the obfuscation model on two parsers. The first parser is used during the training
of our model. This is the bi-affine dependency parser developed by \citet{DBLP:journals/corr/DozatM16}. To test whether our obfuscation model also generalizes to syntactic parsers that were not used during its training, the constituency parser that is included in the AllenNLP software package \cite{gardner2018allennlp} was used.\footnote{We used version 0.8.1.}

For our dependency parser, we follow the canonical setting of using pre-trained word embedding, 1D convolutional character level embedding and POS tag embedding, each of 100 dimensions as the input feature. We also use a three-layer bi-directional LSTM with Bayesian dropout \cite{DBLP:conf/nips/GalG16} as the encoder.
We use the bi-affine attention mechanism to obtain the prediction for each head, and also the prediction for the edge labels.

We use the English Penn Treebank (PTB; \citealt{marcus1993building}) version 3.0 converted using Stanford dependencies for training the dependency parser. We follow the standard parsing split for
training (sections 01--21), development (section 22) and test sets (section 23). The training set portion of the PTB data is also used to train our neural obfuscator model.


We also create a spectrum over the POS tags to decide on the set $\mathcal{P}$ for each of our experiments (see Section~\ref{section:main-model}). This spectrum is described in Table~\ref{table:obf}.
Let the $i$th set in that table be $\mathcal{P}_i$ for $i \in [5]$\footnote{For an integer $k$, we denote by \([k]\) the set \(\{1, . . . , k\}\).}. In our $j$th experiment, $j \in [5]$, we obfuscate
the set $\mathcal{P} = \cup_{i=1}^j \mathcal{P}_i$. This spectrum of POS tags describes a range from words that are highly content-bearing
for privacy concerns (such as named entities) to words that are less of a privacy concern (such as adverbs).

\begin{table}[t!]
\begin{center}
\begin{tabular}{|l|l|l|}
\hline 
$i$ & Category description & $\mathcal{P}_i$ \\
\hline
1 & Named entities & NNP, NNPS \\
2 & Nouns & NN, NNS \\
3 & Adjectives & JJ, JJR, JJS \\
4 & Verbs & VB, VBN, VBD, \\  & & VBZ, VBP, VBG \\
5 & Adverbs & RB, RBR, RBS \\
\hline
\end{tabular}
\end{center}
\caption{\label{table:obf} A spectrum of part-of-speech tags to obfuscate. In the $j$th experiment, we set $\mathcal{P} = \cup_{i=1}^j \mathcal{P}_i$.}
\end{table}

We compare our model against a (privacy) upper-bound baseline which is found to be rather strong. With this baseline, a word $x$ with a tag $t \in \mathcal{P}$ is substituted
with another by a word that appeared with the same tag in the training data from the set $\mathcal{V}_t$. The substituted words are uniformly sampled.
This random baseline serves as an {\em upper} bound for the privacy level achieved, not a lower bound. Randomly substituting a word with another makes
it difficult to recover the original word. However, in terms of parsing accuracy, as we see below, there is a significant room for improvement over that
baseline. There are words, which when substituted by other group of words, yield altogether better parsing accuracy.

\ignore{
 94.1 / 68.3  = 1.37774524158126
  94.1 / 66.9  = 1.40657698056801
 94.3 / 68.4  = 1.37865497076023
 94.3 / 66.4  = 1.42018072289157

  93.7 / 70.7  = 1.32531824611033
  93.7 / 70.3  = 1.33285917496444
 94.1 / 69.7  = 1.35007173601148
 94.1 / 69.4  = 1.35590778097983

  93.1 / 71.9  = 1.29485396383866
  93.1 / 72.3  = 1.28769017980636
 93.6 / 70.5  = 1.32765957446809
 93.6 / 70.1  = 1.33523537803138

  85.2 / 68.1  = 1.2511013215859
  85.2 / 80.2  = 1.06234413965087
 87.3 / 65.3  = 1.33690658499234
 87.3 / 78.1  = 1.11779769526248

  86.4 / 67.2  = 1.28571428571429
  86.4 / 81.2  = 1.064039408867
 88.6 / 64.2  = 1.38006230529595
 88.6 / 77.5  = 1.14322580645161

}

\begin{table*}[th!]

{\small
(a) 
\begin{center}
\begin{tabular}{|l|l|cccccc|cccccc|}
\hline
&		 & \multicolumn{6}{c|}{Random (baseline)}                          & \multicolumn{6}{c|}{Neural model} \\
& Obf. terms &     &  & \multicolumn{2}{c|}{trained}     & \multicolumn{2}{c|}{BERT}        &   &  & \multicolumn{2}{c|}{trained}   & \multicolumn{2}{c|}{BERT} \\
\hhline{~~------------}
 &                &  \multicolumn{2}{c|}{acc (U L)}   & prv & ratio & prv & ratio &  \multicolumn{2}{c|}{acc (U L)}  & prv & ratio & prv & ratio \\
\hline
\multirow{6}{*}{\rotatebox{90}{trained dep.}}  &  Named ent.  &   94.1  &  93.0  &  68.3  & 2.97 &  66.9  & 2.84 &  94.3  &  92.9  &  68.4  & 2.98 &  66.4  & 2.81\\ 
 &  +Nouns           &   93.7  &  92.9  &  70.7  & 3.20 &  70.3  & 3.15 &  94.1  &  92.4  &  69.7  & 3.11 &  69.4  & 3.08\\ 
 &  $\,$$\,$+Adjectives      &   93.1  &  92.4  &  71.9  & 3.31 &  72.3  & 3.36 &  93.6  &  91.7  &  70.5  & 3.17 &  70.1  & 3.13\\ 
 &  $\,$$\,$$\,$$\,$+Verbs           &   85.2  &  80.4  &  68.1  & 2.67 &  80.2  & 4.30 &  87.3  &  78.7  &  65.3  & 2.52 &  78.1  & 3.99\\ 
 &  $\,$$\,$$\,$$\,$$\,$$\,$+Adverbs         &   86.4  &  78.7  &  67.2  & 2.63 &  81.2  & 4.60 &  88.6  &  76.6  &  64.2  & 2.47 &  77.5  & 3.94\\ 
\hhline{~-------------}
& No obf. & \multicolumn{12}{c|}{95.0/93.5 (U/L)} \\
\hline
\hline
\multirow{6}{*}{\rotatebox{90}{AllenNLP dep.}}  &  Named ent.  &   91.9  &  89.7  &  68.3  & 2.90 &  66.9  & 2.78 &  92.2  &  90.1  &  68.4  & 2.92 &  66.4  & 2.74\\ 
 &  +Nouns           &   91.5  &  89.2  &  70.7  & 3.12 &  70.3  & 3.08 &  91.5  &  89.4  &  69.7  & 3.02 &  69.4  & 2.99\\ 
 &  $\,$$\,$+Adjectives      &   90.8  &  88.5  &  71.9  & 3.23 &  72.3  & 3.28 &  91.2  &  89.0  &  70.5  & 3.09 &  70.1  & 3.05\\ 
 &   $\,$$\,$$\,$$\,$+Verbs           &   78.2  &  75.3  &  68.1  & 2.45 &  80.2  & 3.95 &  82.2  &  79.4  &  65.3  & 2.37 &  78.1  & 3.75\\ 
 &   $\,$$\,$$\,$$\,$$\,$$\,$+Adverbs         &   76.7  &  73.5  &  67.2  & 2.34 &  81.2  & 4.08 &  82.0  &  78.9  &  64.2  & 2.29 &  77.5  & 3.64\\ 
\hhline{~-------------}
& No obf. & \multicolumn{12}{c|}{94.2/92.6 (U/L)} \\
\hline
\end{tabular}
\end{center}

(b)
\begin{center}
\begin{tabular}{|l|l|ccccc|ccccc|}
\hline
&		 & \multicolumn{5}{c|}{Random (baseline)}                          & \multicolumn{5}{c|}{Neural model} \\
& Obf. terms &     & \multicolumn{2}{c|}{trained}     & \multicolumn{2}{c|}{BERT}        &    & \multicolumn{2}{c|}{trained}   & \multicolumn{2}{c|}{BERT} \\
\hhline{~~----------}
 &                &  acc (F$_1$) & prv & ratio & prv & ratio &  acc (F$_1$) & prv & ratio & prv & ratio \\
\hline
\multirow{6}{*}{\rotatebox{90}{AllenNLP const.}}  &  Named ent.  &   92.4     &  68.3  & 2.91 &  66.9  & 2.79 &  92.5      &  68.4  & 2.93 &  66.4  & 2.75\\
 &  +Nouns           &   88.2       &  70.1  & 2.95 &  70.3  & 2.97 &  89.0      &  69.7  & 2.94 &  69.4  & 2.91\\
 &   $\,$$\,$+Adjectives      &   86.8       &  71.9  & 3.09 &  72.3  & 3.13 &  88.1      &  70.5  & 2.99 &  70.1  & 2.95\\
 &   $\,$$\,$$\,$$\,$+Verbs           &   79.2       &  68.1  & 2.48 &  80.2  & 4.00 &  82.5      &  65.3  & 2.38 &  78.1  & 3.77\\
 &   $\,$$\,$$\,$$\,$$\,$$\,$+Adverbs         &   76.8       &  67.2  & 2.34 &   81.2  & 4.09 &  79.5      &  64.2  & 2.22 &  77.5  & 3.53\\
\hhline{~-----------}
& No obf. & \multicolumn{10}{c|}{93.7} \\
\hline
\end{tabular}
\end{center}

}

\caption{(a) Results of parsing accuracy and attacker error for two different dependency parsers. ``acc'' denotes accuracy (Unlabeled Attachment Score/Labeled Attachment Score for
the dependency parsers), ``prv'' denotes the attacker error (trained attacker and
BERT attacker as described in Section~\ref{section:ff-attack} Section~\ref{section:bert-attack}) and ``ratio'' is the ratio between
the parser accuracy and the attacker error. Two parsers are considered: a parser that participates in the obfuscation
model optimization (top part), and offline-trained parsers
from the AllenNLP for dependency (bottom part). Two obfuscation models are considered:
neural (Section~\ref{section:main-model}) and a random baseline. ``No obf.'' are parsing results without obfuscation.
See Table~\ref{table:obf} for a description of each category of obfuscation terms.\label{table:results}. Note that the categories are expanded in the cumulative fashion: e.g., ``+Adjectives'' refers to the union of named entities, nouns and adjectives. ``acc'' and ``prv'' are better when they are higher.
(b) Results of parsing accuracy and attacker error for the AllenNLP constituency parser.   ``acc'' denotes accuracy (F$_1$ PARSEVAL). The constituency parser does not participate in the obfuscation model optimization.
\emph{The results demonstrate how quickly the parsers degrade when more terms obfuscated with the random baseline, while retaining a much higher accuracy with the neural system (acc. column).}
}
\end{table*}

We evaluate the model through three different principal metrics. The first metric is the accuracy of the parser (Unlabeled Attachment Score or Labeled Attachment Score, depending on the context\footnote{The unlabeled/labeled distinction refers to whether the label of the edge in the dependency tree is taken into account when computing the accuracy.}),
the second is the mean reciprocal rank (MRR) of the attacker on the relevant method. With each of the attacker methods, we obtain a distribution $q$ over the possible true outputs recovered from the obfuscated word. We use this distribution to calculate the MRR, the average inverse rank of the correct word according to $q$. More precisely, the score we calculate
is: 

\begin{equation}
\mathrm{MRR}={\frac{100}{N}}\sum_{i=1}^{N}{\frac{1}{r_i}},
\end{equation}

\noindent where $r_i \in \mathbb{N}$ is the rank of the $i$th word (in the whole corpus) according to $q$ (the distribution over possible output words
for that word).\footnote{Note that we have a multiplier of $100$ in our $\mathrm{MRR}$ score definition. This deviates from the standard definition of this score.}
The result we report is attacker error, or $100-\mathrm{MRR}$ (the higher it is, the more privacy is maintained).
Finally, we also report the ratio between the accuracy of the parser\footnote{The accuracy is labeled attachment score in the case of dependency parsing.} and the accuracy of an attacker ($\mathrm{MRR}$).
This metric provides a way to measure the amount of accuracy we gain for each point of privacy we lose, in the form of $\displaystyle\frac{\text{accuracy}}{\text{breach}}$
units.

All neural experiments were run on a single GeForce GTX 1080 Ti GPU. The time to run each of the experiments was in the range of 13.3 hours to 25.2 hours.




\begin{table*}[t]

\begin{center}

\begin{minipage}{0.8\textwidth}
\begin{tabular}{l|lllllll}
original & I   &\emph{do}         & \emph{n't}     & \emph{feel}   & \emph{very}      & \emph{ferocious}       &. \\
random & I   &\emph{liberalize} & \emph{Usually} & \emph{spin}   & \emph{firsthand} & \emph{undistinguished} &. \\
neural & I   &\emph{have}       &\emph{not}     &\emph{choose} &\emph{even}      &\emph{Preliminary}     &. \\
POS & PRP &VBP        &RB      &VB     &RB        &JJ              &.
\end{tabular}

$\,$ \\

\begin{tabular}{l|llllllll}
original & \emph{Individuals} &can &\emph{always}   &\emph{have}      &their &\emph{hands}    &\emph{slapped} &. \\
random & \emph{drugstores}  &can &\emph{secretly} &\emph{galvanize} &their &\emph{persons}  &\emph{hurt}    &. \\ 
neural & \emph{brokerages}  &can &\emph{even}     &\emph{get}       &their &\emph{Outflows} &\emph{vetoed}  &. \\
POS & NNS         &MD & RB       &VB        &PRP\$  &NNS      &VBN     &. 
\end{tabular}

$\,$ \\

\begin{tabular}{l|llllllll}
original & \emph{Analysts}   & \emph{do}    & \emph{n't}          & \emph{see}       & it  & that & \emph{way}       & . \\ 
random & \emph{carpenters} & \emph{merge} & \emph{unilaterally} & \emph{undertake} & it  & that & \emph{wind}      & . \\
neural & \emph{brokerages} & \emph{have}  & \emph{not}          & \emph{choose}    & it  & that & \emph{direction} & . \\
POS & NNS  &      VBP &  RB   &      VB   &     PRP & DT &  NN    &    . 
\end{tabular}

$\,$ \\

\begin{tabular}{l|lllll}
original & The &\emph{device}  &   \emph{was}   & \emph{replaced} &.  \\
random &  The& \emph{admiral}  &  \emph{echoed} & \emph{blunted}  &.   \\
neural & The &\emph{insulation}& \emph{were} &    \emph{vetoed}   &.    \\
POS & DT&  NN     &    VBD  &  VBN      & . 
\end{tabular}

$\,$ \\

\begin{tabular}{l|llllllll}
original & `` & That & \emph{was}    &  \emph{offset} & by & \emph{strength}   &    \emph{elsewhere}  & .   \\
random &   `` & That & \emph{produced} & \emph{flawed} & by & \emph{professionalism} & \emph{near}  &     .    \\
neural & `` & That & \emph{were}   &  \emph{vetoed} & by & \emph{direction} &       \emph{even}   &      .   \\
POS & `` & DT  & VBD   &   VBN  &  IN   &  NN   &     RB        & . 

\end{tabular}

\end{minipage}

\end{center}

\ignore{
[____pos] _ROOT_POS NNS        VBP   RB           VB        PRP DT   NN        . 
[____ori] _ROOT     Analysts   do    n't          see       it  that way       . 
[rnd_obf] _ROOT     carpenters merge unilaterally undertake it  that wind      . 
[ctx_obf] _ROOT     brokerages have  not          choose    it  that direction . 

all changed except for "it that"

[____pos] _ROOT_POS `` DT   VBD      VBN    IN NN              RB        . 
[____ori] _ROOT     `` That was      offset by strength        elsewhere . 
[rnd_obf] _ROOT     `` That produced flawed by professionalism near      . 
[ctx_obf] _ROOT     `` That were     vetoed by direction       even      .

all except for "That" and "by"

all except for "The"

}

\caption{\label{table:examples}Example of five sentences obfuscated with the random and neural models. Words in italics are the ones being substituted (or the substitutes).
The obfuscated terms are named entities, nouns, adjectives, verbs and adverbs.}
\end{table*}

\subsection{Results with Dependency Parsing}
\label{section:results-dep}




In our first set of experiments, we compare the random substitution model to the neural model of obfuscation on the dependency parsing model that was used
during training. As is seen in Table~\ref{table:results}(a),
both the neural model and the random model, with their substitutions, do not significantly harm the accuracy of the parser for the first three experiments
(corresponds to rows in the table). However, as the number of obfuscated words increases (with the categories of verbs and adverbs), the accuracy of the parser
significantly drops,  significantly more for the random model than the neural model.

The attacker error is in the range of the 70\%--80\% for all experiments. As expected, the random model,
given its inherent unpredictability, makes it more difficult for the attacker to identify
the original words. That leads to the ratio between accuracy and intrrusion level often being better with the random model. In general, it also seems that the BERT attacker gives similar results to the trained attacker for the random baseline, and worse results with the neural model. Finally, it is evident that as we obfuscate more terms, the attacker's accuracy
decreases, with the BERT attacker consistently outperforming the trained attacker.

We next turn to inspect the problem of dependency parsing without a parser that was trained with the neural obfuscation model (bottom part of Table~\ref{table:results}(a)).
We see similar trends there as well, in which the first three experiments give a reasonable performance for both the neural and the random model with a significant
drop in performance for the two experiments that follow. We also see that the differences between the neural obfuscation model and the random model are smaller (though
still significant), pointing to the importance of using the dependency model during the training of the neural model.

\subsection{Results with Constituency Parsing}
\label{section:constituency}

Table~\ref{table:results}(b) describes the results for constituency parsing with the AllenNLP constituency parser as described
in Section~\ref{section:setup}.
The results point to a similar direction as was described for dependency parsing. While the ratio between accuracy and privacy
is slightly better for the random model, there is a significant drop in performance for the fourth and fifth experiments when comparing
the random model to the neural model.

\ignore{
\begin{table}[t]
  \centering
  \begin{tabular}{|l|r|l|}
  \hline \textbf{$n$} & $enc$-neural & $enc$-random \\
\hline
1	& 94.27\%	93.07\%  & 94.13\%	92.89\% \\
\hline
2	& 94.12\%	92.88\%  & 93.76\%	92.43\% \\
\hline
3	& 93.65\%	92.38\%  & 93.10\%	91.75\% \\
\hline
4	& 87.36\%	85.20\%  & 80.39\%	78.70\% \\
\hline
5	& 88.63\%	86.39\%  & 78.67\%	76.64\% \\
\hline
  \end{tabular}
  \caption{Comparison of the $enc$-random and $enc$-neural model}
  \label{res-table}
\end{table}

From this table we could see that our $enc$-neural will constantly outperform the $enc$-random model in terms of dependency parsing, we'll then analyze what do these $enc$-neural models have learned, and how well they preserve the privacy.
}

\subsection{Analysis of Syntactic Preservation}

Table~\ref{table:examples} presents five sentences and their obfuscated versions both by the neural model and the random model. In general, when we inspected
the results for the two models, we found that the neural model tends to replace words by others that have a functional syntactic role that is closer to the original.
For example, in the examples we present, \emph{was} is replaced with \emph{were} and \emph{n't} is replaced with \emph{not}. The random model, however, does not adhere
to any syntactic similarity between the original word and its substituted version beyond them having been seen in the training data with the same part-of-speech tag.

To further test whether the neural model preserves other syntactic similarities between the original and obfuscated sentences, we
took all verbs from Propbank \cite{kingsbury2002treebank} and created a signature for each one: the list of argument types 
it can appear with.
For example, the signature for \emph{yield}
is \emph{01,012}, which means that ``yield'' appears with two frames in Propbank, one with two arguments and the other with three arguments.
We then calculated for each verb\footnote{The verbs were lemmatized first using the WordNet lemmatizer available in NLTK.}
that appears in the original sentence the overlap between its signature and the signature of the verb in the obfuscated sentence (neural or random).
This overlap is counted as the size of the intersection of the frame signatures of the two verbs. For example, the signature of
\textit{advocate} might be \emph{012} while the signature of \textit{affect} is \textit{012,01}. Therefore, their overlap is $1$.

There was a stark difference between the two averages of the overlap sizes. For the random baseline model, the average was 1.46 (over 5,680 tokens) and for the neural model
the average was 1.80. The difference between these two averages is statistically significant with $p$-value $< 0.05$ in a one-sided $t$-test.

\ignore{

\subsection{Analysis}

To further see if the $enc$-neural model learned to generate syntactically consistent substitution rather than only overfits the pre-trained dependency parser it is trained with, we use another parser that is not presented during the training of the $enc$-neural model to evaluate the two kind of models. 

Shay: were these experiments already done?

Shay: how does the random baseline work?

}



\section{Related Work}

There has been a significant increase in interest in the topic of privacy in the NLP community in recent years. For example, \newcite{DBLP:conf/acl-nlpcss/ReddyK16} focused on obfuscation of gender features from social media text, while \newcite{li2018towards}, \newcite{DBLP:conf/emnlp/CoavouxNC18} and \newcite{DBLP:conf/emnlp/ElazarG18} focused on the removal of private information from neural representations such as named entities and demographic information. Unlike the latter work, we are interested in preserving the privacy of the {\em inputs} themselves, while requiring no extra work from deployed NLP software which processes these inputs. \newcite{marujo2015privacy}, for example, perform multi-document summarization on an approximate version of the original documents.

\ignore{
Another actively researched field is differential privacy \cite{DBLP:conf/tamc/Dwork08}, which  also regards data privacy protection and data processing elsewhere. The main purpose of differential privacy is to enable distribution of the data as a training dataset while at the same time protecting individuals from being identified based on their records in the dataset. There is also recent research that brings differential privacy into natural language processing such as \cite{fernandes2019generalised}, which targets the removal of authorship identity using  differential privacy and the bag-of-words privacy mechanism.
}

Differential privacy \cite{DBLP:conf/tamc/Dwork08} which aims to protect the privacy of information contained in a dataset has also been actively researched. Recent research brings differential privacy into natural language processing, for example, the work by \citet{fernandes2019generalised} that targets the removal of authorship identity in a text classification dataset. 

With homomorphic encryption being a longstanding important topic in cryptography, it has also made its way into the field of privacy in machine learning, particularly in terms of designing neural networks which enable homomorphic operations over encrypted data \cite{DBLP:journals/corr/abs-1711-05189,DBLP:conf/crypto/BourseMMP18}. For example, \newcite{DBLP:conf/icml/Gilad-BachrachD16} designed a fully homomorphic encrypted convolutional neural network that was able to solve the MNIST dataset with practical efficiency and accuracy.
The scheme of direct homomorphic encryption \cite{DBLP:journals/toct/BrakerskiGV14} is constrained by the multiplication depth degree in the circuit, which makes deep models intractable. Other schemes were developed in recent years \cite{DBLP:conf/asiacrypt/CheonKKS17,DBLP:journals/iacr/FanV12,DBLP:journals/corr/abs-1810-00845}, but achieving satisfactory performance is still a challenge. To the best of our knowledge, no prior work has demonstrated that homomorphic encryption could be directly applied to the design of recurrent neural networks or discrete tokens as input.

\section{Conclusions}

We presented a model and an empirical study for obfuscating sentences so that the obfuscated sentences transfer syntactic information from the
original sentence. Our neural model outperforms in parsing accuracy a strong random baseline when many of the words in the sentence
are obfuscated. In addition, the neural model tends to replace words in the original sentence with words which have a closer syntactic function to the original word
than a random baseline.

\section*{Acknowledgments}

The authors thank Marco Damonte and the anonymous reviewers for feedback and comments on a draft of this paper.
This research was supported by a grant from Bloomberg, an ERC Starting Grant BroadSem 678254 and the
Dutch National Science Foundation NWO VIDI grant 639.022.518.

\bibliography{acl2019}
\bibliographystyle{acl_natbib}

\end{document}